\documentclass[sigconf]{aamas}  

\usepackage{booktabs}
\usepackage{graphicx}
\usepackage{subcaption}
\usepackage{balance}

\setcopyright{ifaamas}  
\acmDOI{}  
\acmISBN{}  
\acmConference[AAMAS'18]{Proc.\@ of the 17th International Conference on Autonomous Agents and Multiagent Systems (AAMAS 2018)}{July 10--15, 2018}{Stockholm, Sweden}{M.~Dastani, G.~Sukthankar, E.~Andr\'{e}, S.~Koenig (eds.)}  
\acmYear{2018}  
\copyrightyear{2018}  
\acmPrice{}  

\begin{document}

\title{Discovering Blind Spots in Reinforcement Learning}  




%
\author{Ramya Ramakrishnan}
\affiliation{
  \institution{Massachusetts Institute of Technology}
}
\email{ramyaram@mit.edu}

\author{Ece Kamar}
\affiliation{
  \institution{Microsoft Research}
}
\email{eckamar@microsoft.com}

\author{Debadeepta Dey}
\affiliation{
\institution{Microsoft Research}
}
\email{dedey@microsoft.com}

\author{Julie Shah}
\affiliation{%
  \institution{Massachusetts Institute of Technology}
}
\email{julie_a_shah@csail.mit.edu}

\author{Eric Horvitz} 
\affiliation{%
  \institution{Microsoft Research}
}
\email{horvitz@microsoft.com}

\begin{abstract}
Agents trained in simulation may make errors in the real world due to mismatches between training and execution environments. These mistakes can be dangerous and difficult to discover because the agent cannot predict them a priori. We propose using oracle feedback to learn a predictive model of these \textit{blind spots} to reduce costly errors in real-world applications. We focus on blind spots in reinforcement learning (RL) that occur due to incomplete state representation: The agent does not have the appropriate features to represent the true state of the world and thus cannot distinguish among numerous states. We formalize the problem of discovering blind spots in RL as a noisy supervised learning problem with class imbalance. We learn models to predict blind spots in unseen regions of the state space by combining techniques for label aggregation, calibration, and supervised learning. The models take into consideration noise emerging from different forms of oracle feedback, including demonstrations and corrections. We evaluate our approach on two domains and show that it achieves higher predictive performance than baseline methods, and that the learned model can be used to selectively query an oracle at execution time to prevent errors. We also empirically analyze the biases of various feedback types and how they influence the discovery of blind spots.
\end{abstract}

%


\maketitle


\section{Introduction}

Agents designed to act in the open world are often trained in a simulated environment to learn a policy that can be transferred to a real-world setting. Training in simulation can provide experiences at low cost, but mismatches between the simulator and world can degrade the performance of the learned policy in the open world and may lead to costly errors. For example, consider a simulator for automated driving that includes components to learn how to drive, take turns, stop appropriately, etc. The simulator does not, however, include any emergency vehicles, such as ambulances or fire trucks. When an agent trained in such a simulator is in the proximity of an emergency vehicle in the open world, it will confidently keep driving rather than pull over because it has no knowledge of the mismatch, potentially leading to costly delays and accidents.

\begin{figure}
\label{fig:problem}
\includegraphics[width=0.25\textwidth]{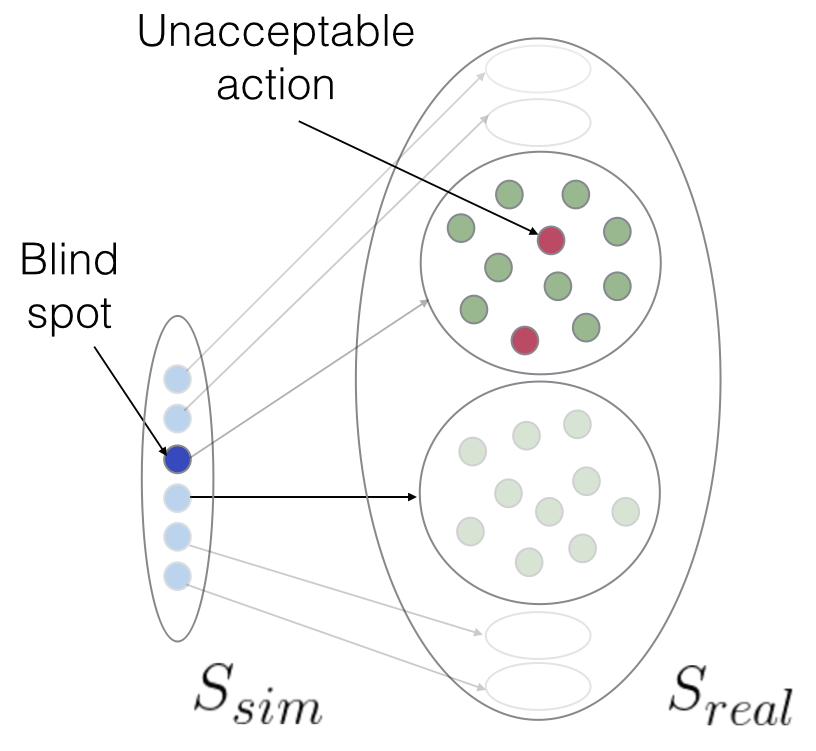}
\caption{Mismatch between simulator state representation ($S_{sim}$) and open-world state representation ($S_{real}$), which can lead to blind spots.}
\label{fig:state_mismatch}
\end{figure}

We first formally define the problem of discovering blind spots in reinforcement learning (RL). \emph{Blind spots} are regions of the state space where agents make unexpected errors because of mismatches between a training environment and the real world. Different kinds of limitations lead to different types of blind spots. We focus in this paper on blind spots that stem from limitations in state representation.  
Limitations in the fidelity of the state space result in the agent being unable to distinguish among different real-world states, as highlighted in Figure \ref{fig:state_mismatch}. In the driving example, for an agent trained in the simulator, states with and without an ambulance appear the same according to the learned representation. However, the optimal action to take in the open world for these two situations differs significantly, and it is impossible for an agent that cannot distinguish these states from one another to learn to act optimally, regardless of the quantity of simulation-based training experience.

Such representational incompleteness is ubiquitous in any safety-critical RL application, especially in robotics since real-world data can be dangerous to collect. An expert could make the agent's simulator representation more complete if the true representation was known a priori, but even with extensive engineering, there is often a gap between simulation and reality. Further, the agent's representation cannot be pre-specified when it is learned automatically through deep RL. In cases where it is impossible to have the complete real-world representation, the agent must first \emph{identify} blind spots, which then enables representation and policy refinement.

We propose a transfer learning approach that uses a learned simulator policy and \emph{limited} oracle feedback to discover such blind spots towards reducing errors in the real world. The oracle provides information by performing the task (demonstrations) or monitoring and correcting the agent as it acts (corrections), which provides signals to the agent of whether its actions were acceptable in each of those states. Our setup is different from prior transfer learning scenarios \cite{taylor2009transfer,barreto2016successor,tobin2017domain,barrett2010transfer,christiano2016transfer} as they do not reason explicitly about the mismatch in state representation between source and target worlds.

We assume that blind spots do not occur at random and correlate with features known to the agent. For example, the agent may lack the feature for recognizing emergency vehicles but the existence of emergency vehicles may correlate with observable features, like vehicle size or color. Under this assumption, we formalize the problem of discovering blind spots as a supervised learning problem, where the objective is to learn a blind spot map that provides the likelihood of each simulation state being a blind spot for the agent by generalizing the observations to unseen parts of the simulation space. That is, the agent learns for every simulation state, the probability that it corresponds to at least one real-world state in which the agent's chosen action will lead to a high-cost error.

We note that learning a predictive model for blind spots is preferred over updating a learned policy when the agent's state representation is insufficient. 
In the driving example, two states that are indistinguishable to the agent require different actions: a state with an ambulance requires pulling over to the side and stopping, while a state without one requires driving at the speed limit. If the agent updates its policy for these similarly appearing states, the consequence can be costly and dangerous. 
Instead, a blind spot model can be used in any safety-critical real-world setting where the agent can prompt for help in potentially dangerous states instead of incorrectly committing to a catastrophic action. 

Formalizing blind spot discovery as a supervised learning problem introduces several challenges: (1) Each observation from the oracle provides a noisy signal about whether the corresponding simulation state is a blind spot. Since a simulation state may correspond to multiple real-world states, identifying whether a simulation state is a blind spot requires aggregating multiple observations together. In addition, the accuracy of observations varies across different types of oracle feedback. For example, corrections clearly indicate whether an agent's action is acceptable in a state, whereas demonstrations only show when agent and oracle behaviors differ, (2) blind spots can be rare and thus learning about them is an imbalanced learning problem, and (3) oracle feedback collected through executions (corrections and demonstrations) violates the i.i.d. assumption and introduces biases in learning. Our approach leverages multiple techniques to address the noise and imbalance problems. Prior to learning, we apply expectation maximization (EM) to the dataset of oracle feedback to estimate noise in the observations and to reduce noise through label aggregation. We apply oversampling and calibration techniques to address class imbalance. Finally, we experiment with different forms of oracle feedback, including random observations, corrections, and demonstrations, to quantify the biases in different conditions.

We evaluate our approach on two game domains. The results show that blind spots can be learned with limited oracle feedback more effectively than baseline approaches, highlighting the benefit of reasoning about different forms of feedback noise. Further, the learned blind spot models are useful in selectively querying for oracle help during real-world execution. Evaluations also show that each feedback type introduces some bias that influences the blind spots that the agent learns. Overall, corrections are informative as they provide direct feedback on the agent's actions. The effectiveness of demonstrations varies: in some cases, demonstrations do not cover important errors the agent may make, resulting in inadequate coverage of all blind spots, while in other cases, demonstration data is sufficient for the agent to avoid dangerous regions altogether.

Our contributions are four-fold: (1) formalizing the problem of discovering blind spots in reinforcement learning due to representation incompleteness, (2) introducing a transfer learning framework that leverages human feedback to learn a blind spot map of the target world, 3) evaluating our approach on two simulated domains, and 4) assessing the biases of different types of human feedback.

\section{Problem Formulation}

We formulate the problem of discovering blind spots in reinforcement learning as a transfer learning problem with a source, or simulated, environment $M_{sim}$ and a target, or real-world, environment $M_{real}$. We assume access to a simulator for $M_{sim}$, in which the agent simulates and learns an optimal policy $\pi_{sim}$. Our goal is to use this learned policy and a limited budget $B$ of feedback from an oracle $O$ to learn a blind spot model of the target world, which indicates the probability that each simulator state is a blind spot. This learned model can then be used to query a human for help at simulator states with a high probability of being a blind spot. Prior work \cite{argall2009survey,knox2009interactively,saunders2017trial,christiano2017deep,griffith2013policy} has investigated the use of human feedback for guiding agents in RL tasks, but we use oracle feedback to learn a blind spot model of the target world rather than to learn a policy.

In our problem, the simulator and real-world environments are mismatched: $M_{sim} \neq M_{real}$. Specifically, the state representation of $M_{sim}$ is limited because observable features of states in the real world are missing. This results in the agent seeing significantly different states of the real world as the same, as shown in Figure \ref{fig:state_mismatch}. Formally, if the source task has state space $S_{sim} = \{ s^1_{sim},...,s^m_{sim} \}$ and the target has state space $S_{real} = \{ s^1_{real},...,s^n_{real} \}$, many real-world states map to each simulator state $\{ s^j_{real},...,s^k_{real} \} \mapsto s^i_{sim}$, where $\{j,...,k\} \in [1,n], \forall i \in [1,m]$. The agent can only reason in the simulator state space $S_{sim}$ because these are the only states it can observe, while the oracle has access to the true real-world state space and provides feedback through $S_{real}$.

An oracle $O = \{A(s,a), \pi_{real}\}$ is defined by two components: an acceptable function $A(s,a)$ and an optimal real-world policy $\pi_{real}$. The acceptable function $A(s,a)$ provides direct feedback on the agent's actions by returning 0 if action $a$ is acceptable in state $s$ and 1 otherwise. In our experiments, we simulated $A(s,a)$ by defining acceptable actions as those with values within some $\delta$ of the optimal action value for that state, but $A(s,a)$ can be defined in many ways. The optimal policy $\pi_{real}$ is used when the oracle is providing demonstrations in the real world. In practice, oracles can be humans or other agents with more expensive and/or complementary mode sensors (e.g. lidar cannot see color but can sense 3D shape while cameras can detect color but not 3D shape).

A state in the simulator world $s_{sim} \in S_{sim}$ is defined to be a blind spot $(B(s_{sim}) = 1)$ if
\begin{equation}
\exists s_{real} \in S_{real} \text{  s.t.  } A(s_{real},\pi_{sim}(s_{sim})) = 1.
\end{equation}
In other words, blind spots are states in the simulator in which the agent's learned action is unacceptable in at least one real-world state that maps to it. Intuitively, if two states look identical to the agent, and it is taking an unacceptable action in either of them, the agent should mark this as a blind spot.

The agent's objective is to use the learned policy $\pi_{sim}$ and a limited budget of $B$ labels from the oracle $O = \{A(s,a),\pi_{real}\}$ to learn a blind spot model $M = \{C, t\}$. The classifier $C: S_{sim} \rightarrow \Pr(B(S_{sim}) = 1)$ predicts for each simulator state $s_{sim} \in S_{sim}$, the probability that $s_{sim}$ is a blind spot in the target world, and the probability threshold $t$ specifies the cutoff for classifying a simulator state as a blind spot.

\subsection{Agent Observations}

A perfect observation for learning the blind spot map would be the pair $<s^i_{sim}, l^i_p>$ such that $l^i_p$ is 1 when $ s^i_{sim}$ is a blind spot -- there exists a real world state corresponding to $ s^i_{sim}$ where $\pi_{sim}( s^i_{sim})$ is not acceptable -- and $l^i_p$ is 0 otherwise. Since the oracle and agent have different representations, the oracle can provide observations over $S_{real}$ not $S_{sim}$. Thus, they are associated with \textit{State representation (SR) noise}. In addition, some forms of oracle feedback may provide weaker information about the quality of agent actions; instead of informing whether an action is acceptable, it may indicate when agent and oracle actions differ, referred to as \textit{Action mismatch (AM) noise}. We describe both types of noise below in detail. 

\subsubsection{State representation noise}
State representation (SR) noise occurs because the agent and oracle are operating in two different representations. The simulator state representation is limited and has missing features that cause the agent to see many distinct real-world states as the same. When the oracle provides an observation for a real-world state, the agent cannot disambiguate this observation and thus maps it to the corresponding simulation state, resulting in state representation noise.  

Let $ s^i_{real}$ be a real-world state and $ s^i_{sim}$ be the simulation state $s^i_{real}$ corresponds to. An observation from the perspective of the oracle is defined as a tuple $<s^i_{real}, l^i_a>$, where $l^i_a \in \{0,1\}$ is the resulting label such that  $l^i_a = A(s^i_{real},a^i_{sim})$ and $a^i_{sim} = \pi_{sim}(s^i_{sim})$. However, due to a limited state representation, the observation from the agent's perspective is $<s^i_{sim}, l^i_a>$.  When many real-world states map to the same simulator state in this way, the agent receives many noisy labels for each state. For example, if the agent takes an acceptable action in one real-world state and an unacceptable action in another real-world state that looks the same to the agent, it will receive two labels for this state, one 0 and one 1.

Returning to our definition of blind spots, if the agent receives even one unacceptable label for any real-world state, the corresponding simulator state is automatically a blind spot. Thus, receiving a blind spot observation is a perfect signal that the corresponding simulation state is a blind spot. However, if the agent receives many acceptable labels, the agent cannot mark the state as safe (i.e., not a blind spot) because there may be other real-world states that map to this one where the agent's action will be unacceptable. The main property of SR noise is that receiving many safe labels does not guarantee that the corresponding simulation state is safe.

\subsubsection{Action mismatch noise}
The formulation described so far assumes that the oracle is giving feedback directly on the agent's actions using the acceptable function $A(s,a)$. Another form of feedback is to allow the oracle to provide demonstrations using $\pi_{real}$ while the agent simply observes, which might be lower cost to the oracle than directly monitoring and correcting the agent's actions. In this case, when feedback is not given directly on the agent's actions, an additional form of noise is introduced: action mismatch (AM) noise. If the oracle takes action $a_i$ and the agent planned to take $a_j$, this could be indicative of a blind spot because the agent is not following the optimal action. However, two actions can both be acceptable in a state, so a mismatch does not necessarily imply that the state is a blind spot. The main property of AM noise is that noisy blind spot labels are given for safe states, and the agent should reason about this noise to avoid being overly conservative and labeling many safe states as blind spots. 

\section{Approach}

\begin{figure}[t]
\begin{center}
\includegraphics[width=8.5cm]{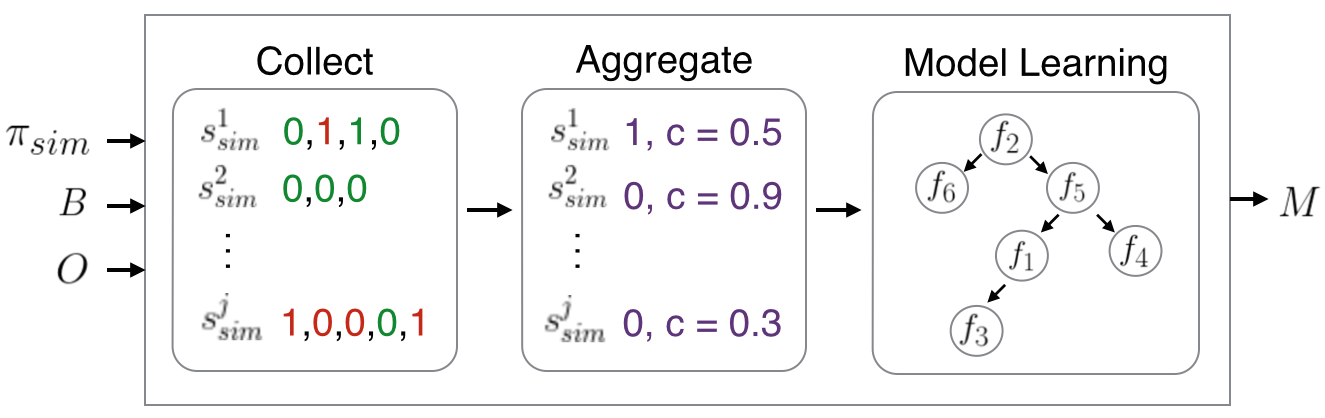}
\end{center}
\caption{Full pipeline of approach.}
\label{fig:pipeline}
\end{figure}

We now present a framework for learning blind spots in RL, shown in Figure \ref{fig:pipeline}. The pipeline includes a data collection phase, in which the agent gets data from an oracle through various forms of feedback. Since each feedback type is associated with noise, we introduce a label aggregation step, which estimates the noise in the labels using EM and predicts the true label of each visited simulation state through aggregation of observations. To generalize observations for visited simulation states to unseen states, we perform supervised learning. The learning step makes the assumption that blind spots do not occur at random and instead are correlated with existing features that the agent has access to.

Since blind spots are often rare in data, learning about them is an imbalanced learning problem. To address this, we first oversample the observations of blind spots and then perform calibration to correct estimates of the likelihood of blind spots. Calibrated estimates are important for our domain as they can be used to decide whether to request oracle help and accurately trade off the likelihood of error with the cost of querying the oracle in execution.    

\subsection{Data collection}

To learn blind spots in the target world, we first need to collect data from an oracle. The oracle can either provide feedback directly on the agent's actions through the acceptable function $A(s,a)$ or demonstrate the optimal action using $\pi_{real}$, which the agent can observe. We now discuss specific forms of oracle feedback and the type of noise (SR and/or AM) that they each have.

We assume that many real-world states look identical to the agent, which results in SR noise. The process for data collection is as follows: the oracle selects a state $s_{real}$ in the target world, the chosen state $s_{real}$ is mapped to $s_{sim}$ in the agent's representation. Then, the agent gets a feedback label $l=\{0,1\}$, where 0 denotes an acceptable action and 1 denotes unacceptable. This label is obtained differently depending on the feedback type. Because many target states map to the same simulator state, the resulting dataset $D = \{(s_{sim}, [l_1,...,l_k])\}$ has many noisy labels for each simulator state.

\noindent \textbf{Random-acceptable (R-A)}: For the first feedback type, states are randomly chosen, and the oracle provides direct feedback on the agent's actions using the acceptable function. In this condition, when an unacceptable label is given for any real-world state, the matching simulator state is marked as a blind spot. In other words, if $\exists l_i = 1, i = [1,..,k]$, $s_{sim}$ is a blind spot. If all labels are 0, or acceptable, the agent can reason about whether states are more or less likely to be blind spots based on the number of 0 labels.

\noindent \textbf{Random-action mismatch (R-AM)}: If the agent could only observe the oracle's action instead of getting feedback on its own action, AM noise is introduced because the agent does not know if taking a different action from the oracle is indicative of an unacceptable action. For this condition, the oracle chooses a state $s_{real}$ which maps to $s_{sim}$, and the agent observes the oracle's action $\pi_{real}(s_{real})$. The agent associates this observation with a noisy unacceptable label if $\pi_{sim}(s_{sim}) \neq \pi_{real}(s_{real})$. The dataset is structured similarly to R-A, with a list of labels for each state, but now there are noisy 1 labels due to action mismatches that are not truly unacceptable. Adding both SR and AM noises make it difficult to recover the true labels, blind spot or not, of these simulator states.

The first two conditions, however, are not realistic, as a person would not give feedback at randomly chosen states. A more natural way to learn from a person is to collect feedback while acting in the real world \cite{christiano2017deep}. Thus, we consider feedback in the form of trajectories $< s^1_{sim}, s^2_{sim}, ..., s^t_{sim} >$ and evaluate the oracle function $A(s,a)$ for each state in the trajectory. We analyze two types of feedback: \emph{demonstrations}, which are optimal trajectories in the real world, and \emph{corrections}, in which the agent acts in the real world while being monitored and corrected by an oracle. Traditional machine learning approaches assume that data is i.i.d., but these feedback types introduce additional biases because states are correlated.

\begin{table}
\centering
\caption{Noise/bias in each feedback type.}
\begin{tabular}{c rrr}
Feedback Type & SR Noise & AM Noise & Bias\\ [0.5ex]
\hline
Random-acceptable (R-A) & \checkmark & - & -\\
Random-action mismatch (R-AM) & \checkmark & \checkmark & -\\
Demo-acceptable (D-A) & \checkmark & - & \checkmark\\
Demo-action mismatch (D-AM) & \checkmark & \checkmark & \checkmark\\
Corrections (C) & \checkmark & - & \checkmark\\
\hline
\end{tabular}
\label{table:feedback}
\end{table}

\noindent \textbf{Demo-action mismatch (D-AM)}: A demonstration is a full trajectory of a task $t_{d} = \{s_0,a_0...,a_{n-1},s_n\}$ from a start state $s_0$ to a goal state $s_n$. The agent obtains a set of states in this trajectory $H = \{ s^i_{sim} | s^i_{sim} \in t_{d}, \pi_{sim}(s^i_{sim}) \neq t_{d}(s^i_{real})\}$, in which the agent's learned source policy action does not match the oracle's action in the demonstration. Similar to the random-action mismatch condition, this action mismatch does not necessarily mean that the agent performed an unacceptable action, as there may be many acceptable actions at that state. However, the agent notes a noisy 1 label for all states with action mismatches $s_{sim} \in H$. This condition, results in a dataset $D = \{(s_{sim}, [l_1,l_2,...,l_k])\}$, with both noisy safe and noisy blind spot labels. 

\noindent \textbf{Demo-acceptable (D-A)}: This feedback is collected similarly to D-AM, followed by a review period for getting direct feedback on the agent's mismatched actions. For all states $s_{sim} \in H$, the agent queries the oracle function $A(s_{real},\pi_{sim}(s_{sim}))$ to resolve all action mismatch ambiguities. Thus, all noisy 1 labels either become a safe label because the agent's action, while different from the oracle's, is actually acceptable or a true blind spot label, which confirms that the agent's action is unacceptable. While AM noise is resolved in this condition, SR noise still exists in all feedback types.

\noindent \textbf{Corrections (C)}: The final feedback we consider is correction data. In this condition, the agent performs one trajectory of a task $t_{c} = \{s_0,a_0...,a_{n-1},s_n\}$, given a start state $s_0$, with the oracle monitoring. If the oracle observes an unacceptable action at any state $s_i \in t_{c}$, it stops the agent and specifies an alternative action. The agent then proceeds until interrupted again or until the task is completed. From this feedback type, we obtain a similar dataset, with no AM noise because the oracle is directly providing feedback on the agent's actions. In this feedback type, queries to $A(s,a)$ are implicit, as an interruption is interpreted as an unacceptable action at that state, and no feedback means that the agent's action is acceptable. 

D-AM is the most difficult type of feedback to reason about because it has SR noise, AM noise, and bias, while corrections is the most informative feedback because it provides direct feedback on the agent's policy. However, we expect demonstrations to be easier to obtain than corrections. In Section \ref{sec:results}, we discuss the biases and tradeoffs of using demonstrations versus corrections data. Table \ref{table:feedback} summarizes all feedback types and the noise/bias found in each.

\subsection{Aggregating Noisy Labels}
For all conditions, the collected data has many noisy labels that need to be aggregated. From the data collection, we obtain a dataset of noisy labels $D_n = \{(s_{sim}, [l_1,l_2,...,l_k])\}$ for states the agent has seen. These labels are noisy due to SR and AM noise. We need to reason about these noisy labels from the oracle to determine what the true label of the state is: blind spot or safe.

For label aggregation, we use the Dawid-Skene algorithm (DS) \cite{dawid1979maximum}, which is a popular approach for addressing label noise in data collection. We prefer the approach since it has a small number of parameters to estimate, it works well over sparse data sets, and it has been shown to consistently work well across problems \cite{sheshadri2013square}. 

DS takes as input a dataset of noisy labels. The goal of DS is to predict the true labels of instances by estimating the prior distribution of the data (ratio of blind spot vs. safe states) and the confusion matrix, which is the noise model between observations (blind spot vs. safe labels) and true labels (blind spot vs. safe states). The algorithm is unsupervised; it uses EM in its estimation. First, it initializes true labels of instances by averaging the observations in each state. Then, it estimates the confusion matrix based on initialized true labels. It uses the confusion matrix to re-estimate the labels by weighting each observation based on the estimated noise. The algorithm iterates until convergence. 

Different forms of feedback are associated with particular noise types that can be used to inform the aggregation approach. For feedback types with no AM noise, there are no noisy blind spot signals for safe states. Thus, we can modify DS to take advantage of this information. We can constrain the top row of the confusion matrix to be [1,0] because we know that safe states will not get any 1 observations. For feedback types with AM noise, there is no structure in the data to constrain the noise estimation problem. Thus, we use the original DS algorithm to learn all parameters. The output of aggregation is a dataset $D_a = \{(s_{sim}, \hat{l}, c)\}$, where $\hat{l}$ is the estimated true label and $c$ is the associated confidence.

\subsection{Model Learning}

With estimated true labels, we train a supervised learner to predict which states are likely to be blind spots in the target world. Because of the relative rarity of blind spot states, the major challenge in model learning is class imbalance. With only a few blind spots, the model will learn to predict all states as safe, which can be extremely dangerous. To deal with class imbalance, we oversample blind spot states to get balanced classes in the training data and then calibrate the model to provide better estimates.

The full model learning process is as follows: A random forest (RF) classifier is trained with data from aggregation $D_a = \{(s_{sim}, \hat{l}, c)\}$, with states being weighted according to $c$. The output is a blind spot model $M = \{C, t\}$, which includes a classifier $C$ and a threshold $t$. To learn $M$, we perform a randomized hyperparameter search over RF parameters, and for each parameter configuration, we run three-fold cross-validation with oversampled data and obtain an average F1-score. We choose the hyperparameters with the highest average F1-score to train the final model.

For the final training, we reserve 30\% of the full training data for calibration. We oversample the rest of the data and train an RF classifier using the best parameters. For calibrating the model after training on oversampled data, we vary the threshold that specifies a cutoff probability for classifying a state as a blind spot. For each possible threshold $t$, we measure the percentage of blind spots predicted by the model on the held-out calibration data and choose $t$ such that the prior of blind spots on the calibration set matches the prior in the training data. The final output $M = \{C, t\}$ is the learned RF classifier $C$ and the threshold $t$.

\section{Experimental Setup}

We conducted experiments in two domains. For each domain, we used one version to train the agent and a modified version to simulate a real-world setting that does not match the training scenario.

\subsection{Domains}
 
The first domain is a modified version of the game Catcher, in which the agent must catch falling fruits. A fruit starts from the top at a random $x$ location and falls straight down at constant speed. The agent controls a paddle at the bottom of the screen that can stay still, move left, or move right. The state of the game is represented as $[x_p, x_f, y_f]$, where $x_p$ is the $x$ location of the player and ($x_f$, $y_f$) represents the fruit's location. In the source task, reward is proportional to the player's $x$ distance away from the fruit, or $W - |x_p - x_f|$, where $W$ is the width of the screen. The target task is split into two regions: the left-hand side, which looks exactly like the source task and the right-hand side, which is like the source with probability $p$ and with probability $1-p$, a ``bad" fruit, instead of the original fruit, falls. For bad fruits, the agent gets higher reward for moving away from it, denoted by $|x_p - x_f|$. An additional high negative reward, -100, is given when $x_p = x_f$ because being right under the fruit is a high danger region. The agent does not have the fruit type feature in its representation, so it does not have the ``true" representation of the real world. Without being able to distinguish the fruit type, the agent can never learn the optimal policy.

The second domain is a variation of FlappyBird. The goal is to fly a bird through the space between two pipes. The state is represented by $[y_t, y_b, y_a, v_a, \Delta x]$, where $y_t$, $y_b$, and $y_a$ are the $y$ locations of the top pipe, bottom pipe, and agent respectively, $v_a$ is the agent's velocity, and $\Delta x$ is the $x$ distance between the agent and the pipe. The agent can either go up or take no action, in which case gravity starts pulling it down. The source task has high pipes and low pipes, and the agent must learn to fly high above the ground and swoop down to make it into both low and high pipes. The agent receives +10 for getting past a pipe, -10 for crashing, and +0.1 any time it flies above a certain threshold (to encourage flying high). In the target task, pipes are made of different materials, copper and steel, which the agent cannot observe. Copper pipes close to the ground can cause heavy wind to pass through, so the agent should be cautious and fly low, but for steel pipes, the agent should continue to fly high. The reward function stays the same for the target task, except that for the special copper pipes, the agent receives +0.1 for flying below a specific threshold (to encourage flying low) and -100 when it flies high because this is a high danger region for copper pipes. Without knowledge of the pipe's material, the agent cannot learn the optimal policy for both types of pipes. 

\subsection{Oracle Simulation}

We assume an oracle $O = \{A(s,a), \pi_{real}\}$ that provides feedback to the agent. We simulate an oracle by learning an optimal policy $\pi_{real}$ in the target task and by constructing an acceptable function $A(s,a)$ that specifies which actions are acceptable in each state. This function depends on the domain as well as how strict or lenient an oracle is. A strict oracle may only consider optimal actions as acceptable, while a lenient oracle may accept most actions, except those that lead to significantly lower values. 

To simulate different acceptable functions, we first trained an agent on the true target environment to obtain the optimal target Q-value function $Q_{real}$. Then, we computed for each state $s_{real} \in S_{real}$, the difference in Q-values between the optimal action and every other action $\Delta Q^i_{s_{real}} = Q_{real}(s_{real},a^*) - Q_{real}(s_{real},a_i), \forall a_i \in A$. The set of all Q-value deltas $\{\Delta Q^i_{s_{real}}\}$ quantifies all possible mistakes the agent could make. The deltas are sorted in ascending order from least dangerous mistakes to costly mistakes, and a cutoff delta value $\delta$ is obtained by choosing a percentile $p$ at which to separate acceptable and unacceptable actions.

This cutoff value is used to define the acceptable function in an experimental setting, and consequently the set of blind spots in the task, which are simulator states with at least one unacceptable action in a real-world state mapping to it. When $A(s,\hat{a})$ is queried with agent action $\hat{a}$, the oracle computes the difference $\Delta Q_{\hat{a}} = Q_{real}(s,a^*) - Q_{real}(s,\hat{a})$. If $\Delta Q_{\hat{a}} < \delta$, action $\hat{a}$ is acceptable in state $s$; otherwise, the action is unacceptable. An acceptable function $A(s,a) = \{Q_{real}, p\}$ is defined by the target Q-value function $Q_{real}$ and a percentile $p$ for choosing the cutoff. A high $p$ value simulates a lenient oracle, resulting in more acceptable actions and fewer blind spot states. Consequently, there is more AM noise because if the oracle accepts most actions, there is a high chance that the agent will take an acceptable action different from the oracle's. A low $p$ value simulates a strict oracle because even actions that have slightly lower Q-values than the optimal will be considered blind spots. This results in less AM noise because deviating from the oracle is a good indicator that the agent's action is truly unacceptable.

\subsection{Baselines}
The first baseline is a majority vote (MV) aggregation method for the noisy labels. For each state, MV takes the label that appears the most frequently as the true label. The second baseline is all labels (AL), which uses no aggregation and simply passes all datapoints to a classifier. The model learning is the same for our method as well as for the baselines. The baselines are used to assess the benefit we get from using DS for aggregation.

We report the performance of baselines in predicting blind spots based on the F1 score to assess both the precision and recall of the predictions as well as the accuracy of estimates of the likelihood of blind spots. We compare results for a strict versus a lenient oracle. The strict oracle was chosen such that only the optimal action was acceptable (no associated percentile $p$), and for the lenient oracle, we used $p=0.95$ for Catcher and $p=0.7$ for FlappyBird.

\section{Results}
\label{sec:results}

We now present results of our approach on both domains. We found that our approach achieves higher performance than existing baselines and that different forms of feedback induce biases in the data that affect the learned blind spot model.

\subsection{Benefits of aggregation}

\begin{figure}
\centering
\begin{minipage}{0.23\textwidth}
\includegraphics[width=\textwidth]{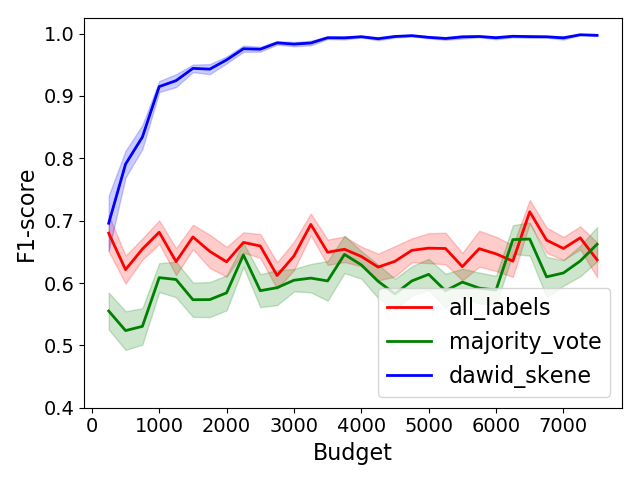}
\subcaption{R-AM, Strict}
\end{minipage}\hfill
\begin{minipage}{0.23\textwidth}
\includegraphics[width=\textwidth]{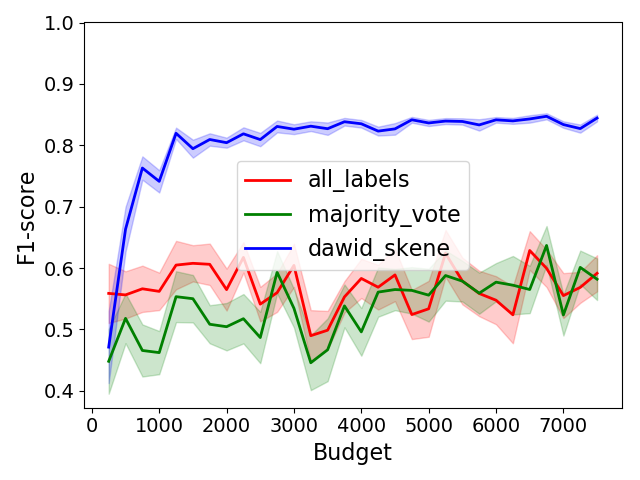}
\subcaption{R-AM, Lenient}
\end{minipage}\hfill
\begin{minipage}{0.23\textwidth}
\includegraphics[width=\textwidth]{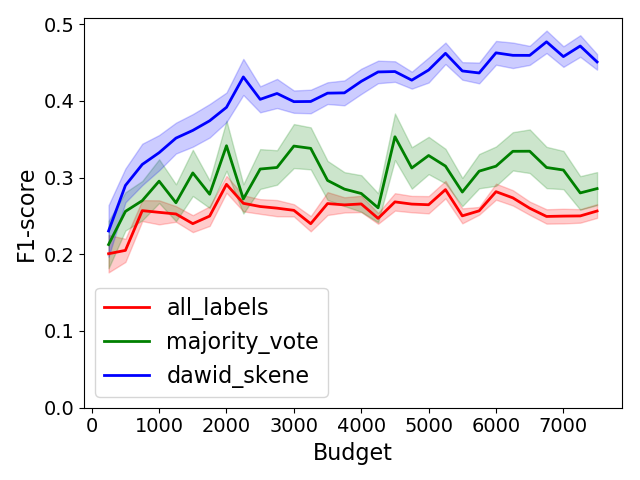}
\subcaption{D-AM, Strict}
\end{minipage}\hfill
\begin{minipage}{0.23\textwidth}
\includegraphics[width=\textwidth]{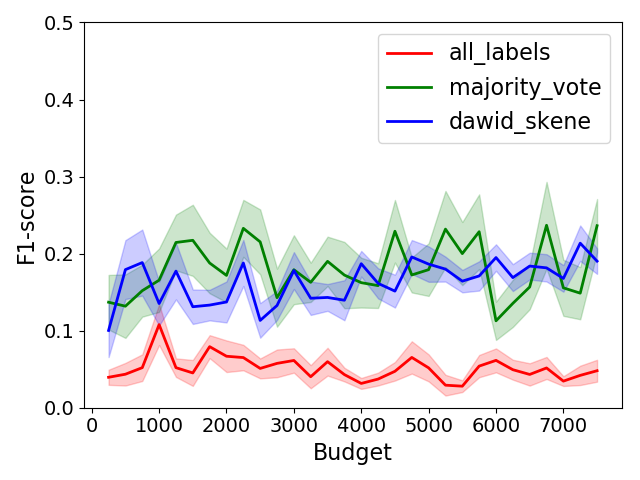}
\subcaption{D-AM, Lenient}
\end{minipage}\hfill
\caption{Comparison of our approach to baseline methods on random and demonstration data with varying oracles.}
\label{fig:agg_results}
\end{figure}

We first compare the performance of our approach, which uses DS for aggregation, to existing baselines -- majority vote (MV) and all labels (AL). In this section, we focus on feedback types with AM noise because DS provides most benefit when the noise cannot be easily recovered by simple techniques. We present results on states seen by the agent during data collection, as this shows the difference between our approach and the baselines on the ability to estimate and reduce noise in the training data. As shown in Figure \ref{fig:agg_results}, we vary the number of oracle labels the agent receives (budget) and report resulting F1-scores, weighted according to the ``importance" of states, which is represented by how often the states are visited by $\pi_{sim}$. We further compare all of the approaches when we have a strict versus a lenient oracle.

For randomly sampled data, using DS performs much better than MV and AL, for both a strict and a lenient oracle because the observations are uniform across all states. DS can thus recover the prior and confusion matrix that generated this data, while MV mostly predicts safe because safe signals are much more common. AL has lower accuracy because it does not aggregate the labels, resulting in a poor prior estimate of blind spot states.

With demonstration data, performance drops overall compared to random data since observations are biased by the oracle's policy, which prevents learning about some blind spots that the agent will face in execution. Despite the decrease in performance, DS still performs well compared to MV and AL for a strict oracle. With a lenient oracle, many safe states are associated with blind spot observations -- due to action mismatch noise -- that an unsupervised learning method like DS cannot recover completely. Nevertheless, DS does equally well to MV and much better than AL.

Overall, DS performs well compared to the baselines. Performance drops when states are sampled in a biased form rather than randomly sampled and when there is a lenient oracle rather than a strict one. We see similar trends for FlappyBird of the benefit of DS over baselines. We discuss details of the effect of feedback types and resulting biases for the two domains in Section 5.3.

\subsection{Effect of feedback type on classifier performance}

Next, we evaluate the best performing approach (learning with DS) as we vary the oracle feedback type. We evaluate the classifier on states seen in oracle feedback, which measures the ability of DS to recover the true labels from noisy state labels, and on unseen data, which highlights the ability of the classifier to generalize to unvisited states. We report F1-scores in Table \ref{table:feedback_types} for each condition. The results show that learning from random data performs well across conditions when the observations have only SR noise (R-A) and both SR and AM noises (R-AM). R-AM can do well in these cases despite AM noise because DS can recover the labels for randomly sampled data when labels are uniformly distributed across states. The performances of both drop when the oracle is lenient, as there are fewer blind spots to learn from.

The results show that the correlated nature of observations from demonstrations and corrections overall reduce the performance of classifiers compared to the random feedback types. 
For the strict oracle case, the performances of D-A, D-AM, and C are comparable, as the AM noise is low. This results in similar feedback from corrections and demonstrations because the monitoring oracle that corrects the agent will redirect the agent any time it deviates from the optimal. 
On the other hand, for a lenient oracle, an oracle correcting an agent will only correct if an action is very dangerous, resulting in a more informative state distribution than demonstrations. 
In this case, both versions of demonstrations fail to collect observations about major blind spot regions. On top of that, D-A ends up having few blind spot observation labels because there is no AM noise. This hurts the prior estimate and makes it hard to learn an accurate classifier. Thus, we see that the performance of D-A is even worse than D-AM for this scenario.

\begin{table}
\centering
\caption{Effect of feedback type on classifier performance in Catcher, reported as F1-scores.}
\begin{tabular}{c |rr|rrr}

 & \textbf{Strict} & & \textbf{Lenient} & \\ [0.5ex]
 \hline
Feedback Type & Seen & Unseen & Seen & Unseen\\ [0.5ex]
\hline
R-A & 0.996 & 0.994 & 0.825 & 0.700 \\
R-AM & 0.997 & 0.993 & 0.837 & 0.833 \\
D-A & 0.476 & 0.453 & 0.084 & 0.152 \\
D-AM & 0.487 & 0.477 & 0.209 & 0.274 \\
C & 0.478 & 0.461 & 0.636 & 0.520 \\
\hline
\end{tabular}
\label{table:feedback_types}
\end{table}

\subsection{Effect of feedback type on oracle-in-the-loop evaluation}

The ultimate goal of our work is to use limited feedback to learn a blind spot model of the target world that can be used to act more intelligently. Ideally, the agent can use this model to selectively query human help to avoid costly mistakes without overburdening the human helper. The next set of results evaluates the effectiveness of the learned model in oracle-in-the-loop (OIL) execution. The agent executes actions in the real-world environment using the source policy. When the learned model predicts a state to be a blind spot using the learned calibrated threshold, the agent queries an oracle for the optimal action. The agent takes this provided action and resumes acting in the world. We compare our method of querying the oracle based on the learned model to an agent that never queries and an overly conservative agent that always queries.

\begin{figure}
\centering
\begin{minipage}{0.17\textwidth}
\includegraphics[width=\textwidth]{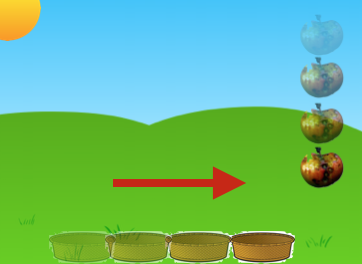}
\subcaption{Catcher, Corrections}
\label{fig:catcher_c}
\end{minipage}\hfill
\begin{minipage}{0.17\textwidth}
\includegraphics[width=\textwidth]{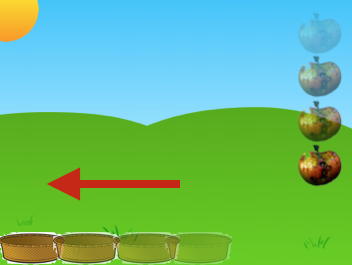}
\subcaption{Catcher, Demo}
\label{fig:catcher_d}
\end{minipage}\hfill
\begin{minipage}{0.17\textwidth}
\includegraphics[width=\textwidth]{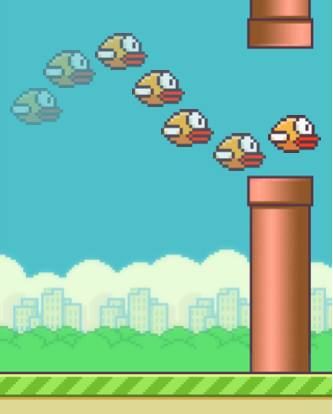}
\subcaption{Flappy, Corrections}
\label{fig:flappy_c}
\end{minipage}\hfill
\begin{minipage}{0.17\textwidth}
\includegraphics[width=\textwidth]{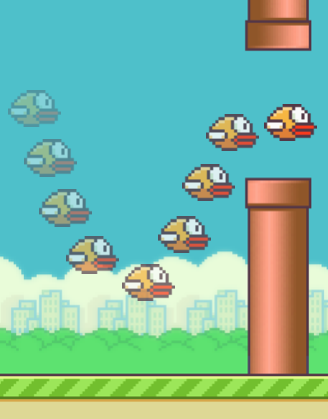}
\subcaption{Flappy, Demo}
\label{fig:flappy_d}
\end{minipage}\hfill
\caption{Data bias of demonstrations and corrections.}
\label{fig:bias}
\end{figure}

Figure \ref{fig:bias} shows that on both domains, demonstrations and corrections provide different feedback and thus introduce separate biases in the data. Corrections give direct feedback on the actions the agent would take, while demonstrations follow the policy of an optimal oracle. In Catcher, an optimal oracle moves towards good fruits and away from bad fruits. An agent trained in the source task with only good fruits learns to move close to all fruits. In Figures \ref{fig:catcher_c} and \ref{fig:catcher_d}, the fruits represent the movement of a bad fruit, and the agent moving at the bottom shows the feedback bias. Demonstrations provide observations about states that are far away from bad fruits, while corrections provide observations about states closer to bad fruits. In FlappyBird, the agent should be careful around copper pipes and fly low. As shown in Figures \ref{fig:flappy_c} and \ref{fig:flappy_d}, a demonstration would show the agent flying low for a copper pipe, while a correction trajectory would allow the agent to fly slightly high and correct only before it goes too far. This provides information about more informative states that the agent will likely visit.

\begin{figure}
\centering
\begin{minipage}{0.23\textwidth}
\includegraphics[width=\textwidth]{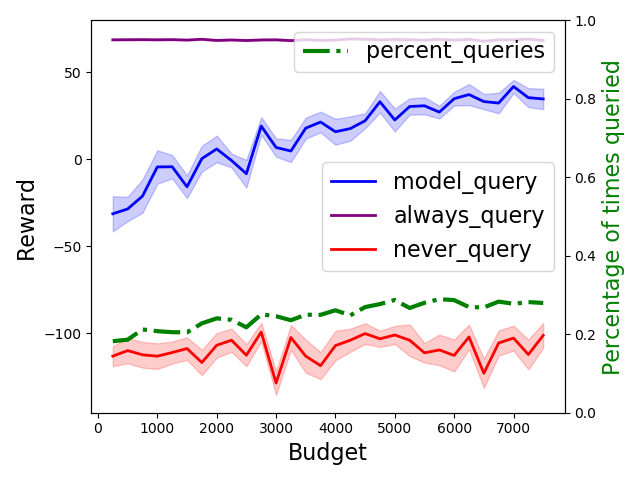}
\subcaption{Lenient, D-AM}
\label{fig:lenient_dam}
\end{minipage}\hfill
\begin{minipage}{0.23\textwidth}
\includegraphics[width=\textwidth]{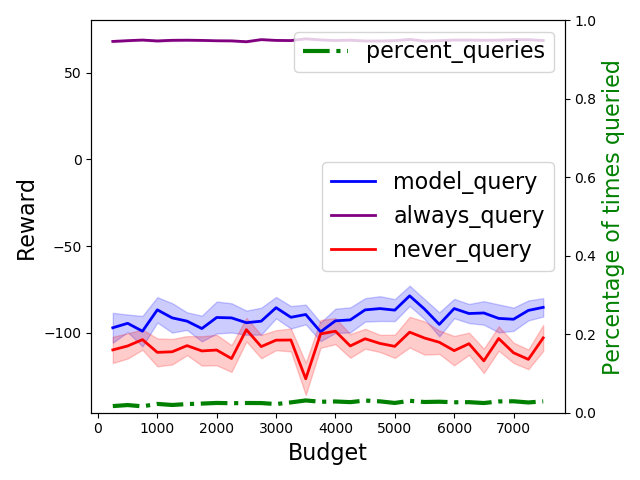}
\subcaption{Lenient, D-A}
\label{fig:lenient_da}
\end{minipage}\hfill
\begin{minipage}{0.23\textwidth}
\includegraphics[width=\textwidth]{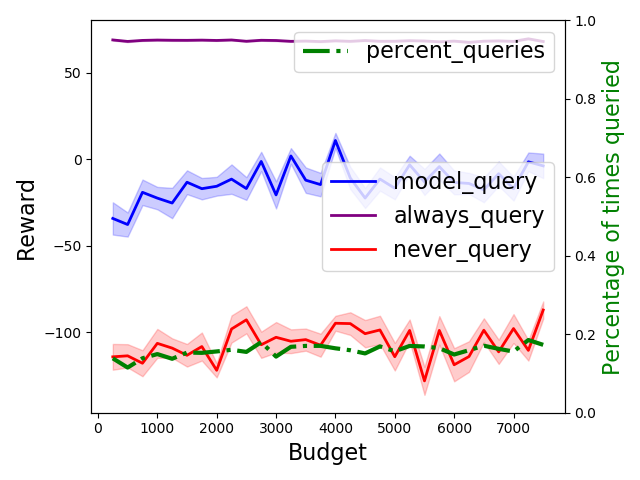}
\subcaption{Lenient, C}
\label{fig:lenient_c}
\end{minipage}\hfill
\begin{minipage}{0.23\textwidth}
\includegraphics[width=\textwidth]{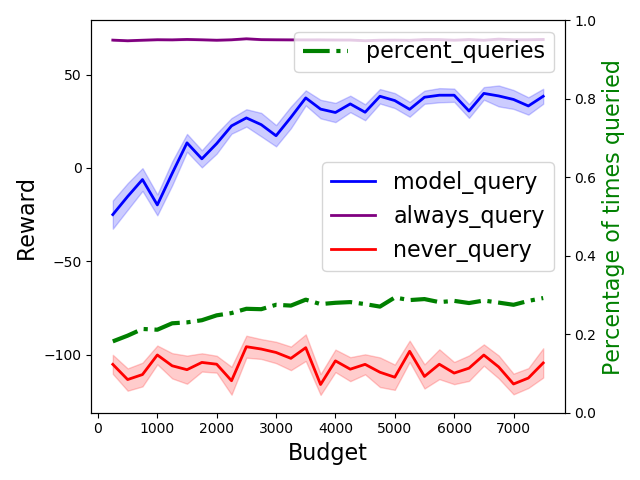}
\subcaption{Strict, D-A}
\label{fig:strict_da}
\end{minipage}\hfill
\caption{Oracle-in-the-loop evaluation on the Catcher domain with varying feedback types.}
\label{fig:hil_eval}
\end{figure}

Figure \ref{fig:hil_eval} shows the performance of our model on OIL evaluation with different feedback types. We report performance as the number of oracle labels (budget) increases. The left y-axis shows the reward obtained on the target task by an agent using the model to query compared to an agent that never queries (NQ) and always queries (AQ). On the same graph, the dotted line indicates the percentage of times the agent queried the human for help using our model. Across all feedback types, we see that the model achieves higher reward than an agent that never queries, while still querying relatively infrequently.

Figures \ref{fig:lenient_dam} and \ref{fig:lenient_c} show that for a lenient oracle, D-AM and C both obtain higher reward than NQ, while C has a lower percentage of queries. Even though D-AM does not get a high F1-score on classifier performance, it does well on OIL evaluation because D-AM considers any action mismatch, where the agent deviates from the optimal, as a blind spot. This results in an overly conservative agent that queries for help at all mismatches. For example, in Catcher, when the oracle is far from a bad fruit and is moving away from it, D-AM marks these states as blind spots due to action mismatches. When the agent queries for help in these states, the oracle instructs the agent to move away. For D-A (Figure \ref{fig:lenient_da}), states far from the fruit are resolved to be safe. The agent thus moves towards it, but because there was never any demonstration data close to bad fruits, D-A has ventured into unknown territory and does not know to act. With a strict oracle (Figure \ref{fig:strict_da}), all states with action mismatches are blind spots, so D-A does not have this issue and queries the oracle in the same states as D-AM.

\begin{table}
\centering
\caption{Reward and percentage of times queried for oracle-in-the-loop evaluation in FlappyBird.}
\begin{tabular}{c |rr|rrr}
 & \textbf{Strict} & & \textbf{Lenient} & \\ [0.5ex]
\hline
Feedback & Reward & \% Queries & Reward & \% Queries\\ [0.5ex]
\hline
AQ & 12.69 & 100\% & 12.66 & 100\%\\
NQ & -318.77 & 0\% & -316.17 & 0\%\\
D-A & -151.15 & 25\% & -373.46 & 5\%\\
D-AM & -197.84 & 20\% & -186.46 & 23\%\\
C & -147.21 & 21\% & -125.44 & 14\%\\
\hline
\end{tabular}
\label{table:flappy}
\end{table}

We see similar trends in FlappyBird, shown in Table \ref{table:flappy}. For a strict oracle, all three feedback types perform similarly and better than NQ, while also querying much less than AQ. For a lenient oracle, D-A performs badly because the initial states, where an oracle demonstrates how to fly low and into the pipe, are resolved to be safe. The agent thus does not query and ends up flying high, resulting in a high negative reward. Note that the classifiers used in OIL evaluation value precision as much as recall, which means it puts equal emphasis on making a mistake and querying the oracle unnecessarily. In cases where errors are more costly than querying, the classifier threshold can be chosen accordingly to increase the aggregate reward of oracle-in-the-loop execution in return for a larger percentage of queries to the oracle. 

\section{Related Work}

\noindent \textbf{Supervised Learning:} In related work, Lakkaraju et al. \cite{lakkaraju2017identifying} introduce a method for finding unknown unknowns in discriminative classifiers. Data points are clustered in an unsupervised manner followed by a multi-arm bandit algorithm (each cluster is an arm) for efficiently finding regions of the feature space where the classifier is most likely to make mistakes. It is not straightforward to apply this approach to RL because examples (states) are no longer i.i.d. as in supervised learning. In RL, states are visited according to a distribution induced by either executing the learned policy or that induced by an optimal oracle. There can also be multiple labels (actions) that are acceptable for each state, rather than a single ``correct" label. Finally, certain mistakes in the real-world can be catastrophic, which requires risk-sensitive classification to prioritize identifying rare blind spot states \cite{zadrozny2003cost}.

\noindent \textbf{Safe Reinforcement Learning:} While RL under safety constraints is an active research topic \cite{garcia2015comprehensive}, many of these techniques \cite{osband2016deep, gal2016dropout, kahn2017uncertainty} are focused on cautious exploration and do not address the scenario where the agent has a flawed state representation, which prevents it from learning calibrated uncertainty estimates.

\noindent \textbf{Novelty/Anomaly Detection:} Anomaly detection \cite{chandola2009anomaly} is related but not directly applicable, as blind spots are not rare instances. Instead, they are regions of the state space where the training environment does not match the testing environment, and we learn to efficiently identify these regions through oracle feedback.

\noindent \textbf{Transfer Learning and Domain Adaptation:} Many approaches improve transfer of information across tasks \cite{taylor2009transfer, pan2010survey,christiano2016transfer,barrett2010transfer}, as tasks cannot be learned from scratch each time. Much of this literature has focused on learning mappings between state and action spaces to enable Q-value function or policy transfer. Several works have also considered hierarchical approaches to RL that involve transferring subtasks across domains \cite{kulkarni2016hierarchical}. In distinction to transfer learning, where the labels (actions) of examples (states) may change, domain adaptation deals with situations where the distribution of examples changes from one domain to another \cite{csurka2017domain,jiang2008literature}. Our work differs from these in that we relax the assumption that the agent's state representation is complete and sufficient to learn autonomously.

\section{Conclusion}

We address the challenge of discovering blind spots in reinforcement learning when the state representation of a simulator used for training is not sufficient to describe the real-world environment. We propose a methodology to explicitly handle noise induced by this representation mismatch as well as noise from low precision oracle feedback. The approach achieves higher performance than baselines on predicting blind spot states in the target environment. We additionally show that this learned model can be used to avoid costly mistakes in the target task while drastically reducing the number of oracle queries. We finally discuss the biases of different types of feedback, namely demonstrations and corrections, and assess the benefits of each based on domain characteristics. Further investigations are needed for ideal integration of blind spot models into oracle-in-the-loop execution by trading off the cost of a mistake with the cost of querying an oracle. In another direction, we see the possibility of moving beyond a heavy reliance on high-quality training data via active learning approaches that can obtain more informative information from the oracle.


\bibliographystyle{ACM-Reference-Format}  
\balance
\bibliography{references}  

\end{document}